\documentclass[10pt, a4paper]{article}
\usepackage{lrec}
\usepackage{multibib}
\newcites{languageresource}{Language Resources}
\usepackage{graphicx}
\usepackage{tabularx}
\usepackage{multirow}
\usepackage{soul}
\usepackage{booktabs}

\usepackage{epstopdf}
\usepackage[utf8]{inputenc}

\usepackage{hyperref}
\usepackage{xstring}

\title{Building a Sentiment Corpus of Tweets in Brazilian Portuguese}

\name{Henrico Bertini Brum, Maria das Gra\c{c}as Volpe Nunes}

\address{Interinstitutional Center for Computational Linguistics (NILC) \\
		 Institute of Mathematical and Computer Sciences, University of S\~{a}o Paulo\\
         henrico.brum@usp.br, gracan@icmc.usp.br\\}

\abstract{
The large amount of data available in social media, forums and websites motivates researches in several areas of Natural Language Processing, such as sentiment analysis. The popularity of the area  due to its subjective and semantic characteristics motivates research on novel methods and approaches for classification. Hence, there is a high demand for datasets on different domains and different languages. This paper introduces TweetSentBR, a sentiment corpora for Brazilian Portuguese manually annotated with $15.000$ sentences on TV show domain. The sentences were labeled in three classes~(positive, neutral and negative) by seven annotators, following literature guidelines for ensuring reliability on the annotation. We also ran baseline experiments on polarity classification using three machine learning methods, reaching $80.99\%$ on F-Measure and $82.06\%$ on accuracy in binary classification, and $59.85\%$ F-Measure and $64.62\%$ on accuracy on three point classification. \\ \newline \Keywords{Sentiment Analysis, Corpus Annotation, Social Media} }

\begin{document}

\maketitleabstract

\section{Introduction} \label{intro}

Sentiment Analysis~(SA) became a popular area of Natural Language Processing in the last decade. The classification of semantic orientation of documents is a challenge for artificial intelligence methods since it is based not only on the regular meaning of words, but also on their semantic role in the context and on the author's intention. Furthermore, the amount of data available in blogs, social media posts and forums has created a great opportunity for researchers to build datasets for evaluating methods and studying new linguistic phenomena.

Websites on e-commerce, movie reviews and hotel reservations usually allow the user to an objective evaluation besides the written commentaries. This objective evaluation~(binary recommendation, star score, 10-point scale) can be a good feature for automatic labeling large datasets on semantic orientation, thus improving the resources for researches over the past decades~\cite{pang2002thumbs,pang2005seeing,blitzer2007biographies}.

The limitation of this technique is the data available in this conditions. Social media, for example, is a large source of user opinions and evaluation~\cite{pak2010twitter}, but the lack of an objective score attached to the posts demands a manual annotation in order to data become useful for SA, even though the data is enriched by linguistic phenomena such as expressions, slangs and irony. 

Manual annotation ends up being more expensive and time consuming, since it demands several guarantees of accuracy, such as developing guidelines, training annotators and revising the annotation~\cite{hovy2010towards}.

In this paper we introduce TweetSentBR~(TTsBR), a corpus manually annotated with data extracted from Twitter. The section \ref{related} presents some related work on SA and corpus annotation. Section \ref{tsbr} presents the corpus and its properties, such as the size, the annotation tags, the information on annotators and the process of data extraction. Section \ref{experiments} presents some data analysis and polarity classification experiments on the corpus. Section \ref{disc} is a brief discussion on the importance of the corpus and how it can be used in Brazilian Portuguese research on SA.

\section{Related Work} \label{related}
Several works present new methods and approaches for tasks such as polarity classification~\cite{turney2002thumbs,pang2005seeing}, detection of irony~\cite{carvalho2009clues,reyes2012humor} and aspect extraction in text~\cite{hu2004mining}.

One of the major issues of this area is the building of datasets for evaluating methods and for training machine learning models. \newcite{turney2002thumbs}, one of the first works on polarity classification, used product reviews labeled as ``recommended'' and ``not recommended''. The source of the data was a website called Epinions, where users could evaluate products and leave a five star score for each review. The authors considered any review with less than 3 stars as ``not recommended''. \newcite{pang2002thumbs} uses a similar score~(star rating) in order to compile a corpus of movie reviews on three classes~(positive, negative and neutral).

The automatic approach worked very well for building large datasets, but the method limited research on domains where users input an objective score. Despite of the challenges of the manual annotation, researches began building new datasets by training annotators to label the data. \newcite{socher2013recursive} introduces Stanford Sentiment Treebank, a re-labeling of the previous IMDB corpus presented in \cite{pang2005seeing}. SemEval, an important semantic evaluation event, also produces several datasets for English designed for SA tasks~\cite{nakov2016semeval}. Some authors even used distant supervision techniques for automatic labeling large datasets quickly using features such as \textit{emoticons}~\cite{go2009twitter}.

In Brazilian Portuguese, several works presented corpora for SA. \newcite{freitas2012vampiro} introduce ReLi, a sentiment corpus of book reviews manually annotated in three classes~(positive, neutral and negative). The authors have chosen books from different publics in order to vary the linguistic phenomena in the corpus~(from teenage books to literature classics). ReLi contains annotation of semantic orientation, part-of-speech tagging and aspect of opinion, and it was later used as resource for researches in SA~\cite{balage2013evaluation,brum2016sentiment}. One of the issues on this corpus observed on the literature is the unbalanced classes - the majority of sentences is neutral~($72\%$), while the negative class represents only $4\%$ of the data.

On the product review domain, \newcite{hartmann2014large} presented Buscape corpus, a large corpora in Brazilian Portuguese. The corpus contains $13.685$ reviews labeled as positive and negative, using scores given by users on Buscape, a popular e-commerce website. A similar dataset is Mercado Livre corpus, introduced in \newcite{avancco2015normalizaccao}, containing $43.818$ product reviews also labeled automatically and balanced between the two classes.

\newcite{silva2011effective} collected a corpus from Twitter in Portuguese. The dataset was collected by searching two entities in the social network~(Dilma and Serra, two running candidates at the time) and manually annotated as positive or negative. The corpus contains $76.358$ documents balanced between positive and negative. The corpus was originally constructed for sentiment stream analysis meaning it contains several retweets and links, phenomena that may interfere on sentiment classification but is vital to maintain the stream for the former task.

Also on binary polarity classification, \newcite{moraes20157x1} introduce the Corpus $7$x$1$, a brazilian portuguese corpus on Twitter comments during the 2014 World Cup semi-finals. The corpus presents some interesting user behavior such as irony, sarcasm, cheering and angry due to the final match score. Corpus $7$x$1$ contains $2.728$ tweets labeled manually in three classes - the neutral class represents tweets that do not align with wither positive or negative sentiments.

\newcite{moraes2016classificaccao} also uses Twitter as the source of data, but compile a corpus of computer products containing $2.317$ tweets. The data is manually labeled in three classes and the authors also performed experiments on SA using lexical-based classifiers and SVM.

A large Twitter corpus was compiled by \cite{correa2017pelesent} using distant supervision. The authors labeled tweets in Brazilian Portuguese using emojis representing positive and negative sentiments following the work of \newcite{go2009twitter} in English. The corpus contains $554.623$ positive tweets and $425.444$ negative. The approach is a fast way to label data, but the method can not guarantee the absence of noise data such as irony, sarcasm or incorrect labels.

\section{TweetSentBR} \label{tsbr}
TweetSentBr is composed of $15.000$ tweets~($17.166$ tokens) extracted using Python-Twitter~\footnote{\url{https://github.com/bear/python-twitter}}, a wrapper for Twitter API. Due to the limitations of Twitter API, we developed a continuous crawler in order to obtain documents during the first semester of 2017. The final dataset is split in two documents - a training set with $12.999$ documents labeled in positive~($44$\%), neutral~($26$\%) and negative~($29$\%); and a test set composed of $2001$ documents with similar distribution to the training set, $45$\%, $25$\% and $29$\% respectively. See \autoref{table:corpus} for the number of documents in each class.

\begin{table}[!h]
\begin{center}
\begin{tabular}{lccc}
\toprule
 \textbf{Class} & \textbf{Training set} & \textbf{Test set} & \textbf{Total} \\ \midrule
      Positive & $5.745$~\textit{($44\%$)} & $903$~($45\%$) & $6.648$\\
      Neutral  & $3.414$~\textit{($26\%$)} & $512$~($25\%$) & $3.926$\\
      Negative & $3.840$~\textit{($29\%$)} & $586$~($29\%$) & $4.426$\\ \midrule
\textbf{Total} & $12.999$         & $2.001$        & $15.000$\\ \bottomrule
\end{tabular}
\caption{Amount of documents in the corpus in each class.} \label{table:corpus}
 \end{center}
\end{table}

\subsection{Data source}
Data was extracted from Twitter between January and July in 2017. We chose to focus on the TV show domain because of the large amount of user generated content on Twitter during the exhibition of the shows. Hashtags~($\#$) are used on social media to group messages on topics and the TV shows usually ask for its audience to use a specific hashtag in order to get visibility in these social networks. Some of the program hashtags group hundreds of thousands of messages during the exhibition of a show and that content can represent suggestions, complaints, evaluations and questions to the entities related to the programs.

We empirically defined nine programs from three major TV channels in Brazil based on its popularity and activeness in social media. Talk-shows, reality shows~(gastronomy and music) and variety shows were chosen in order to diversify the phenomena in the corpus. The periodicity of the exhibitions are also different, some shows go live daily when others go live once or twice a week.

Since we were looking for user generated content, we ignored documents generated by public entities, such as celebrities, companies, TV channels or any official user on Twitter. We also discarded retweets, which are the reposts of popular posts in the social network.

\subsection{Classes definition} \label{class_def}

Following \newcite{hovy2010towards} recommendations, a codebook or manual was written to ensure the agreement between annotators. The codebook contains examples, definitions and tips for the annotation process. The rules and guidelines were formed by empirically observing the dataset crawled before the annotation update based on the feedback from the annotators during the early stages of annotation.

The definitions were created based on the domain and the input received by the annotators after the first contact with the data. These are the guidelines for the annotation in TTsBR:

\textbf{Positive class:} Positive sentences describe feelings of pleasure, satisfaction, compliment or recommendation. The target of the sentiment must be the TV show or any entity related to it~(host, guests, audience, sketches, invited bands...). Positive comparisons, such as ``\textit{This show is better than the other}'' are considered positive and emojis can be strong indicatives of positivity.


\textbf{Negative class:} Negative sentences describe feelings of disagreement, disapprove, complaint or hate. The target of the sentiment must be the TV show or any entity related to it~(hosts, guests, audience, sketches, invited bands...). Negative sentences can be direct, as in ``\textit{Today's show is terrible...}'' or implied in a suggestion, as in ``\textit{the host could improve its jokes, right?}''. Factual information such as delays, abrupt cuts or technical failures are also considered negative as long as it refers for the show or any entity related to it. Emojis are also good indicatives of negativity.


\textbf{Neutral class:} The neutral label must be used for any sentence the annotator could not identify as an opinion~(positive or negative) direct or implied. Factual sentences that do not represent a hit or a flaw, such as ``\textit{Show X just started}'', inaccurate semantic orientation~(``\textit{Don't know what to think about this}'') and sentences the annotator can not completely comprehend were instructed to be annotated as neutral. Some tweets in the corpus were generated by social media robots~(most of them on audience measurements) and the annotators labeled these as neutral as well.


We also wanted to keep track of the sentences that most caused doubt in the annotators. The annotators had a check box to mark in case of doubt in the annotation, even though this option did not prevent the annotator from labeling the sentiment of the sentence. The annotators were instructed to mark the \textit{doubt} option every time they felt divided between two or more classes or when they took more than the average time~($2$ minutes) in a sentence.
The addition of a \textit{doubt} option gives us new information on the data and also reduces the stress on annotators. In the first stages of annotation, only an average of $10\%$ of sentences were marked.

\subsection{Annotation process}
For the annotation process we recruited seven native speakers of Brazilian Portuguese in three different areas - linguistics, journalism and computer science. The annotation process was based on \newcite{hovy2010towards}, following the eight steps of annotation in order to improve the reliability of the resource.

\begin{figure}[!h]
\begin{center}
\includegraphics[scale=0.4]{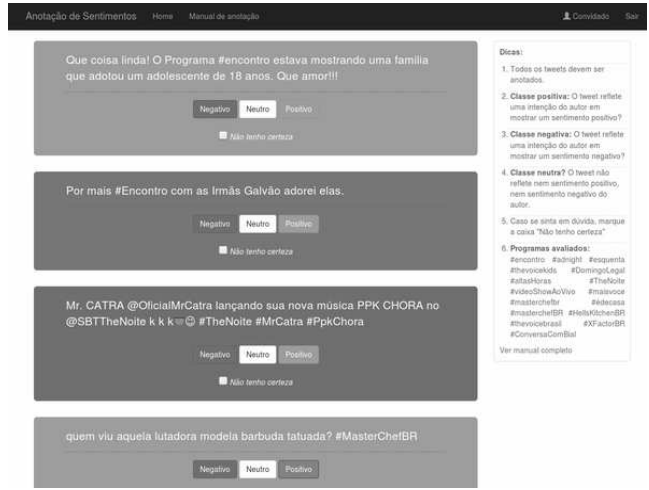} 
\caption{Snapshot of the annotation interface.}
\label{fig:interface}
\end{center}
\end{figure}

We developed a user friendly interface for the annotators to label the tweets~(\autoref{fig:interface}). The interface contains the codebook, the phases of annotation, a progress bar and a panel with tweets for labeling. The annotation panel shows to the annotator the three classes and a box to be checked when in doubt~(even though every tweet must have a label chosen in order to proceed to the next phase) and a side box with quick tips, contact information and a link to the codebook.

Each annotator received a set with one hundred tweets to be labeled. After this step we measured the average time, agreement~(all received the same set) and we took notes of the questions about the codebook guidelines. Then we proceeded to rewrite the codebook, adding more examples and detailing the definitions based on the questions presented on the first annotation.

The next step was a meeting with all annotators to receive the feedback of the annotators, when the tweets were revised by everyone and we presented the new version of the codebook.

We then proceeded to the regular annotation. First the participants labeled $300$ tweets in order to measure agreement. We used Krippendorf's Alpha~\cite{kripendorff2004reliability} to measure the agreement of the annotators. In this phase we obtained $52.9\%$ on the nominal measure and $70\%$ on the interval measure. The annotators began the individual phases when each one labeled around $2.000$ tweets in six weeks, completing the training portion of the corpus.

Two supervisors annotated a small portion of the corpus in order to obtain a test set revised. The goal was to form a $10\%$ part of the dataset specially labeled for evaluating machine learning methods. Some of the data annotated in the agreement phase was also used to compose this set.

For the release, we define the general sentiment for each document based on a major voting of the labels provided by each annotator. Some documents were only labeled by one annotator, while other were annotated by $3$ or $7$ annotators. $45$ documents tied and have no sentiment label, even though they were kept in the dataset with a ``\textit{none}'' label.

\subsection{Release and distribution} \label{distribution}
The dataset is available in \url{http://bitbucket.org/HBrum/tweetsentbr/}. Twitter has a Privacy Policy forbidding the redistribution of the data, so we managed to provide only the \textit{ids} of the tweets in the corpus. Any user with a working identification can search for the tweets freely.

We provide the dataset with the \textit{ids}, the hashtags used in the search, full annotators labels, the count of how many annotators checked the \textit{doubt} option and the general sentiment for each document, as well as a tool for downloading the dataset as long as having a valid credential~(provided by Twitter itself). An example of the dataset is presented below:

\begin{verbatim}
id       hashtag     labels  h  s split
----------------------------------------
86304477	#encontro   [1,1,1] 0  1 train
86558371	#theNoite   [1,0,0] 2  0 test
86506323	#encontro   [1]     1  1 train
86466839	#masterChef [-1]    0 -1 test
\end{verbatim}

\section{Experiments} \label{experiments}

In order to investigate the properties of the corpora, we defined a series of experiments to determine word frequency, the class balance, and we also performed some polarity classification using baseline methods for Portuguese.

For the experiments, we performed a preprocessing of the data - we replaced numbers~(dates, currency values) by a \textit{NUMBER} token, we also replaced user names and links by the tokens \textit{USERNAME} and \textit{URL}, respectively. We trimmed repetition of characters~(eg. ``looooove'' turns into ``love'') to a minimum of $3$ repeat characters

\subsection{Corpus statistics}

In order to extract some information from the corpus, we measured the relevance of words on each class. We calculated the tf-idf value of each term ignoring hashtags, stop-words, emojis and punctuation. We chose to report only the polarity classes~(positive and negative) since the neutral class groups several characteristics~(facts, out-of-topic sentences, confusing content) that led the analysis to terms not expressive, such as the name of the shows, users nicknames and neutral verbs~(present, watch,...).

The terms indicated in \autoref{table:stats_words} show a notable semantic orientation represented in the classes - the positive class shows the verb ``to love'' and positive adjectives, while the negative class shows adjectives~(the word ``trash'' is popular used as adjective on Twitter) and the verb ``to remove'' that may indicate a request for removing a guest from a show or even a participant from a reality show.

\begin{table}[ht]
\centering
\begin{tabular}{lcc|cc}
\toprule
              & \multicolumn{2}{c}{\textbf{Positive class}} & \multicolumn{2}{c}{\textbf{Negative class}} \\ \midrule
\textbf{\#}   & \textbf{PT-BR} & \textbf{EN} & \textbf{PT-BR} & \textbf{EN}  \\ \midrule
1             & amo          & to love       & ridículo & ridiculous  \\
2             & fofura         & cute        & péssimo  & awful       \\
3             & adorando       & loving      & lixo     & trash       \\
4             & emocionada     & emotional   & tirem    & to remove \\
5             & linda          & beautiful   & mala     & boring      \\ \midrule
\end{tabular}
\caption{Five most relevant terms according to tf-idf for each polarity class.}
\label{table:stats_words}
\end{table}

\subsection{Polarity classification task}

In order to evaluate the corpus on the polarity classification task we used three machine learning methods proposed in \newcite{avancco2015normalizaccao} in product review corpora for Brazilian Portuguese. We ran two classic classifiers, Naive Bayes and SVM~(linear kernel), and a hybrid approach initially proposed for cross-domain polarity classification.

The hybrid classifier combines SVM and a lexical-based approach for binary sentiment classification. The lexical-based classifier uses linguistic rules  designed for binary classification, such as identifying strong sentiment words in sentences and increasing or decreasing their value based on the presence of terms~(eg. \textit{very}, \textit{less}, \textit{much}). The lexical-based method is used when the sentence representation is located near the hyperplane~(in a threshold assumed as $0.5$ by the former author). The work has been used in other works in the literature~\cite{avancco2016improving,correa2017pelesent}.

We represented the sentences in vector using a binary bag-of-words~(presence/absence of terms), sentiment words~(obtained through Portuguese lexicons), emoticons and POS tags. The methods were built using Scikit-learn library~\cite{pedregosa2011scikit}.

\autoref{table:results_posneg} presents the results obtained for the binary classification which is a common approach in the area~\cite{turney2002thumbs,avancco2016improving,nakov2016semeval}. The neutral class was not used in this experiments.

\begin{table}[ht]
\centering
\begin{tabular}{lcc}
\toprule
\textbf{Method}  & \textbf{F-Measure}  & \textbf{Accuracy} \\ \midrule 
Naive Bayes      & 0.7987              & 0.8099            \\
SVM              & 0.8099              & 0.8206            \\
Hybrid Classifier& 0.7659              & 0.7684            \\
\bottomrule
\end{tabular}
\caption{Pos/neg classification of TweetSentBR.}
\label{table:results_posneg}
\end{table}

For the three class classification we followed the pipeline approach used in \newcite{moraes2016classificaccao} - first we classify in \textit{objective} and \textit{subjective}~(meaning neutral or polarity sentences, respectively) and then classify the polarity one in positive or negative. We did not evaluate the corpora using the Hybrid Classifier since it uses a lexical-based approach that could not be re-factored to analyze the neutral class. The results are shown in \autoref{table:opinions_three}

\begin{table}[ht]
\centering
\begin{tabular}{lcc}
\toprule
\textbf{Method}  & \textbf{F-Measure}  & \textbf{Accuracy} \\ \midrule 
Naive Bayes      & 0.5985              & 0.6462            \\
SVM              & 0.5964              & 0.6462            \\
\bottomrule
\end{tabular}
\caption{Pos/neu/neg classification of TweetSentBR.}
\label{table:opinions_three}
\end{table}

\section{Discussion and future work} \label{disc}

TweetSentBR is a manually annotated corpus designed for polarity classification. The corpus was formed using a novel domain for the Brazilian Portuguese language what can be exploited by new machine learning approaches such as deep learning architectures and ensembles.

It also offers new resources for linguistic approaches on natural language by observing the expressions, social media behavior or hate speech detection. The \textit{doubt} label, for example, can be used for a better evaluation of classifiers by comparing machine learning flaws with human uncertainty on labeling data.

This corpora also differs from other approaches by including the neutral class. The addition of the neutral class approximates the corpora to popular applications, since the polarity classifiers available in the industry must find solutions for separating the opinions of users from noisy data. ReLi~\cite{freitas2012vampiro} and Computer-BR~\cite{moraes2016classificaccao} are the only corpora we found in the literature that describes the use of a neutral class on the annotation.

We believe this corpora can still be improved by labeling more data manually or by using semi-supervised methods, such as self-training or co-training~\cite{da2016using}.
\section{Acknowledgement} \label{ack}

We acknowledge financial support from CNPq and the help of Amanda Carneiro, Fernando Nóbrega, Juliana Batista, Rafael Anchiêta and Thales Bertaglia. for the volunteer annotation of TTsBR and also Marcos Treviso for the development of the annotation interface.

\section{Bibliographical References}
\label{main:ref}

\bibliographystyle{lrec}
\bibliography{paper}

\begin{thebibliography}{}

\bibitem[\protect\citename{Avan{\c{c}}o \bgroup et al.\egroup
  }2016]{avancco2016improving}
Avan{\c{c}}o, L.~V., Brum, H.~B., and Nunes, M.
\newblock (2016).
\newblock Improving opinion classifiers by combining different methods and
  resources.
\newblock {\em XIII Encontro Nacional de Intelig{\^e}ncia Artificial e
  Computacional (ENIAC)}, pages 25--36.

\bibitem[\protect\citename{Avan{\c{c}}o}2015]{avancco2015normalizaccao}
Avan{\c{c}}o, L.~V.
\newblock (2015).
\newblock Sobre normaliza{\c{c}}{\~a}o e classifica{\c{c}}{\~a}o de polaridade
  de textos opinativos na web.

\bibitem[\protect\citename{Balage \bgroup et al.\egroup
  }2013]{balage2013evaluation}
Balage, P.~P., Pardo, T. A.~S., and Alu{\i}sio, S.~M.
\newblock (2013).
\newblock An evaluation of the brazilian portuguese liwc dictionary for
  sentiment analysis.
\newblock {\em Proceedings of the 9th Brazilian Symposium in Information and
  Human Language Technology (STIL)}, pages 215--219.

\bibitem[\protect\citename{Blitzer \bgroup et al.\egroup
  }2007]{blitzer2007biographies}
Blitzer, J., Dredze, M., Pereira, F., et~al.
\newblock (2007).
\newblock Biographies, bollywood, boom-boxes and blenders: Domain adaptation
  for sentiment classification.
\newblock {\em ACL}, 7:440--447.

\bibitem[\protect\citename{Brum \bgroup et al.\egroup }2016]{brum2016sentiment}
Brum, H., Araujo, F., and Kepler, F.
\newblock (2016).
\newblock Sentiment analysis for brazilian portuguese over a skewed class
  corpora.
\newblock {\em International Conference on Computational Processing of the
  Portuguese Language}, pages 134--138.

\bibitem[\protect\citename{Carvalho \bgroup et al.\egroup
  }2009]{carvalho2009clues}
Carvalho, P., Sarmento, L., Silva, M.~J., and De~Oliveira, E.
\newblock (2009).
\newblock Clues for detecting irony in user-generated contents: oh...!! it's so
  easy;-.
\newblock In {\em Proceedings of the 1st international CIKM workshop on
  Topic-sentiment analysis for mass opinion}, pages 53--56. ACM.

\bibitem[\protect\citename{Correa~Junior \bgroup et al.\egroup
  }2017]{correa2017pelesent}
Correa~Junior, E.~A., Marinho, V.~Q., Santos, L. B.~d., Bertaglia, T.~F.,
  Treviso, M.~V., and Brum, H.~B.
\newblock (2017).
\newblock Pelesent: Cross-domain polarity classification using distant
  supervision.
\newblock {\em arXiv preprint arXiv:1707.02657}.

\bibitem[\protect\citename{da Silva \bgroup et al.\egroup }2016]{da2016using}
da~Silva, N. F.~F., Coletta, L.~F., Hruschka, E.~R., and Hruschka~Jr, E.~R.
\newblock (2016).
\newblock Using unsupervised information to improve semi-supervised tweet
  sentiment classification.
\newblock {\em Information Sciences}, 355:348--365.

\bibitem[\protect\citename{Freitas \bgroup et al.\egroup
  }2012]{freitas2012vampiro}
Freitas, C., Motta, E., Milidi{\'u}, R., and C{\'e}sar, J.
\newblock (2012).
\newblock Vampiro que brilha... r{\'a}! desafios na anota{\c{c}}ao de opiniao
  em um corpus de resenhas de livros.
\newblock {\em Encontro de Ling{\'i}stica de corpus}, 11:3.

\bibitem[\protect\citename{Go \bgroup et al.\egroup }2009]{go2009twitter}
Go, A., Bhayani, R., and Huang, L.
\newblock (2009).
\newblock Twitter sentiment classification using distant supervision.
\newblock {\em CS224N Project Report, Stanford}, 1(2009):12.

\bibitem[\protect\citename{Hartmann \bgroup et al.\egroup
  }2014]{hartmann2014large}
Hartmann, N.~S., Avan{\c{c}}o, L.~V., Balage~Filho, P.~P., Duran, M.~S., Nunes,
  M. d. G.~V., Pardo, T. A.~S., and Aluisio, S.~M.
\newblock (2014).
\newblock A large corpus of product reviews in portuguese: tackling
  out-of-vocabulary words.
\newblock {\em 9th International Conference on Language Resources and
  Evaluation}.

\bibitem[\protect\citename{Hovy and Lavid}2010]{hovy2010towards}
Hovy, E. and Lavid, J.
\newblock (2010).
\newblock Towards a 'science' of corpus annotation: a new methodological
  challenge for corpus linguistics.
\newblock {\em International journal of translation}, 22(1):13--36.

\bibitem[\protect\citename{Hu and Liu}2004]{hu2004mining}
Hu, M. and Liu, B.
\newblock (2004).
\newblock Mining and summarizing customer reviews.
\newblock In {\em Proceedings of the tenth ACM SIGKDD international conference
  on Knowledge discovery and data mining}, pages 168--177. ACM.

\bibitem[\protect\citename{Kripendorff}2004]{kripendorff2004reliability}
Kripendorff, K.
\newblock (2004).
\newblock Reliability in content analysis: Some common misconceptions.
\newblock {\em Human Communications Research}, 30:411--433.

\bibitem[\protect\citename{Liu}2012]{liu2012sentiment}
Liu, B.
\newblock (2012).
\newblock Sentiment analysis and opinion mining.
\newblock {\em Synthesis lectures on human language technologies}, 5(1):1--167.

\bibitem[\protect\citename{Moraes \bgroup et al.\egroup }2015]{moraes20157x1}
Moraes, S. M.~W., Manssour, I.~H., and Silveira, M.~S.
\newblock (2015).
\newblock 7x1pt: um corpus extra{\'\i}do do twitter para an{\'a}lise de
  sentimentos em l{\'\i}ngua portuguesa.
\newblock {\em Proceedings of Symposium in Information and Human Language
  Technology}.

\bibitem[\protect\citename{Moraes \bgroup et al.\egroup
  }2016]{moraes2016classificaccao}
Moraes, S.~M., Santos, A.~L., Redecker, M.~S., Machado, R.~M., and Meneguzzi,
  F.~R.
\newblock (2016).
\newblock Classifica{\c{c}}{\~a}o de sentimentos em n{\'\i}vel de
  senten{\c{c}}a: uma abordagem de m{\'u}ltiplas camadas para em lingua
  portuguesa.
\newblock {\em XIII Encontro Nacional de Inteligência Artificial e
  Computacional}.

\bibitem[\protect\citename{Nakov \bgroup et al.\egroup }2016]{nakov2016semeval}
Nakov, P., Ritter, A., Rosenthal, S., Sebastiani, F., and Stoyanov, V.
\newblock (2016).
\newblock Semeval-2016 task 4: Sentiment analysis in twitter.
\newblock {\em Proceedings of the 10th International Workshop on Semantic
  Evaluation (SemEval 2016)}.

\bibitem[\protect\citename{Pak and Paroubek}2010]{pak2010twitter}
Pak, A. and Paroubek, P.
\newblock (2010).
\newblock Twitter as a corpus for sentiment analysis and opinion mining.
\newblock In {\em LREc}, volume~10.

\bibitem[\protect\citename{Pang and Lee}2005]{pang2005seeing}
Pang, B. and Lee, L.
\newblock (2005).
\newblock Seeing stars: Exploiting class relationships for sentiment
  categorization with respect to rating scales.
\newblock In {\em Proceedings of the 43rd Annual Meeting on Association for
  Computational Linguistics}, ACL '05, pages 115--124, Stroudsburg, PA, USA.
  Association for Computational Linguistics.

\bibitem[\protect\citename{Pang \bgroup et al.\egroup }2002]{pang2002thumbs}
Pang, B., Lee, L., and Vaithyanathan, S.
\newblock (2002).
\newblock Thumbs up?: Sentiment classification using machine learning
  techniques.
\newblock In {\em Proceedings of the ACL-02 Conference on Empirical Methods in
  Natural Language Processing - Volume 10}, EMNLP '02, pages 79--86,
  Stroudsburg, PA, USA. Association for Computational Linguistics.

\bibitem[\protect\citename{Pedregosa \bgroup et al.\egroup
  }2011]{pedregosa2011scikit}
Pedregosa, F., Varoquaux, G., Gramfort, A., Michel, V., Thirion, B., Grisel,
  O., Blondel, M., Prettenhofer, P., Weiss, R., Dubourg, V., et~al.
\newblock (2011).
\newblock Scikit-learn: Machine learning in python.
\newblock {\em Journal of Machine Learning Research}, 12(Oct):2825--2830.

\bibitem[\protect\citename{Reyes \bgroup et al.\egroup }2012]{reyes2012humor}
Reyes, A., Rosso, P., and Buscaldi, D.
\newblock (2012).
\newblock From humor recognition to irony detection: The figurative language of
  social media.
\newblock {\em Data \& Knowledge Engineering}, 74:1--12.

\bibitem[\protect\citename{Silva \bgroup et al.\egroup
  }2011]{silva2011effective}
Silva, I.~S., Gomide, J., Veloso, A., Meira~Jr, W., and Ferreira, R.
\newblock (2011).
\newblock Effective sentiment stream analysis with self-augmenting training and
  demand-driven projection.
\newblock In {\em Proceedings of the 34th international ACM SIGIR conference on
  Research and development in Information Retrieval}, pages 475--484. ACM.

\bibitem[\protect\citename{Socher \bgroup et al.\egroup
  }2013]{socher2013recursive}
Socher, R., Perelygin, A., Wu, J., Chuang, J., Manning, C.~D., Ng, A., and
  Potts, C.
\newblock (2013).
\newblock Recursive deep models for semantic compositionality over a sentiment
  treebank.
\newblock In {\em Proceedings of the 2013 conference on empirical methods in
  natural language processing}, pages 1631--1642.

\bibitem[\protect\citename{Turney}2002]{turney2002thumbs}
Turney, P.~D.
\newblock (2002).
\newblock Thumbs up or thumbs down?: Semantic orientation applied to
  unsupervised classification of reviews.
\newblock In {\em Proceedings of the 40th Annual Meeting on Association for
  Computational Linguistics}, ACL '02, pages 417--424, Stroudsburg, PA, USA.
  Association for Computational Linguistics.

\end{thebibliography}

\end{document}